\documentclass[conference]{IEEEtran}
\IEEEoverridecommandlockouts
\usepackage{cite}
\usepackage{amsmath,amssymb,amsfonts, bbm}
\usepackage{algorithm, algorithmic}
\usepackage{subcaption}
\usepackage{graphicx}
\usepackage{textcomp}

\usepackage{xcolor}
\usepackage{placeins} 
\def\BibTeX{{\rm B\kern-.05em{\sc i\kern-.025em b}\kern-.08em
    T\kern-.1667em\lower.7ex\hbox{E}\kern-.125emX}}

\begin{document}

\title{Real-Time Detection and Analysis of Vehicles and Pedestrians using Deep Learning}

\author{\IEEEauthorblockN{ Md Nahid Sadik}
\IEEEauthorblockA{\textit{Department of Computing Science} \\
\textit{University of Alberta}\\
Edmonton, Canada \\
msadik@ualberta.ca}
\and
\IEEEauthorblockN{ Tahmim Hossain}
\IEEEauthorblockA{\textit{Department of Computing Science} \\
\textit{University of Alberta}\\
Edmonton, Canada \\
tahmim@ualberta.ca}
\and
\IEEEauthorblockN{ Faisal Sayeed}
\IEEEauthorblockA{\textit{Department of Computing Science} \\
\textit{University of Alberta}\\
Edmonton, Canada \\
faisalz1@ualberta.ca}
}

\maketitle

\begin{abstract}
Computer vision, particularly vehicle and pedestrian identification is critical to the evolution of autonomous driving, artificial intelligence, and video surveillance. Current traffic monitoring systems confront major difficulty in recognizing small objects and pedestrians effectively in real-time, posing a serious risk to public safety and contributing to traffic inefficiency. Recognizing these difficulties, our project focuses on the creation and validation of an advanced deep-learning framework capable of processing complex visual input for precise, real-time recognition of cars and people in a variety of environmental situations. On a dataset representing complicated urban settings, we trained and evaluated different versions of the YOLOv8 and RT-DETR models. The YOLOv8 Large version proved to be the most effective, especially in pedestrian recognition, with great precision and robustness. The results, which include Mean Average Precision and recall rates, demonstrate the model's ability to dramatically improve traffic monitoring and safety. This study makes an important addition to real-time, reliable detection in computer vision, establishing new benchmarks for traffic management systems.

\end{abstract}


\section{Introduction}

The advent of self-driving cars and the spread of sophisticated traffic control solutions heralded a new era in transportation technology. The crucial role of Computer Vision (CV) technologies, particularly in the areas of vehicle and pedestrian detection and categorization, is central to these breakthroughs. These technologies are critical not only for improving the efficiency and safety of autonomous vehicles but also for solving the broader difficulties of urban traffic management.

Object identification, a key component of Computer Vision, has found widespread use in autonomous systems for recognizing automobiles, pedestrians, traffic lights, and other vital aspects in a traffic scenario. The combination of these technologies considerably reduces the dangers involved with self-driving vehicles. For example, accurate vehicle identification can help to avoid collisions, lane detection can help to maintain proper driving directions, and detection of traffic signals and signage assures adherence to traffic rules.

Furthermore, traffic management has developed as a serious societal burden in recent years, aggravated by factors such as pollution, parking scarcity, and congestion. Technology advancements, combined with lower production costs, have increased the usage of high-resolution video cameras in surveillance systems. These systems create massive amounts of data that are beyond the reach of manual examination, emphasizing the importance of automated solutions\cite{r0}.

Recognizing and classifying different vehicle kinds is crucial not just for autonomous driving systems but also for monitoring. Vehicle classification is an important feature of traffic management software that necessitates prior knowledge of vehicle kinds and models. This entails comprehensive feature extraction and categorization, with applications spanning a wide spectrum of traffic control applications.

Existing methods, while rapid, frequently lack precision, especially in real-time recognition of small objects and pedestrians. This constraint causes significant safety hazards and inefficiencies in transportation, particularly in metropolitan areas where automobiles and humans encounter regularly and unexpectedly. Accuracy and dependability in a variety of settings, such as congested streets, changing weather, and varying lighting scenarios, are critical to ensuring the safety of all road users.

This project aims to develop a fast and accurate vehicle and pedestrian detection system by incorporating state-of-the-art object detection algorithms. Our objective is to create a deep learning framework specifically designed for this nuanced task, capable of processing complex visual data efficiently and adapting to various environmental conditions. By integrating advanced deep learning techniques, the model will identify both vehicles and pedestrians with high precision in real-time. The implementation of this sophisticated model is expected to significantly enhance safety measures in traffic management and revolutionize how traffic is monitored and controlled, thereby contributing to safer, more intelligent, and more efficient urban mobility. 

\section{Literature Review}
Vehicle detection and classification is mainly an object detection and classification problem. For
object detection, several categories of methods have been used over the years. From manual
sliding window method, to statistical models, to machine learning models to eventually deep
learning models. Deep learning models perform exceedingly well in terms of speed compared
to the other types of models.

Various approaches have been investigated in the attempt to construct sophisticated systems for real-time car and pedestrian identification utilizing deep learning, each bringing unique insights and facing significant problems. This section examines three key techniques in the field: the integration of YOLOv4 with DeepSORT, EnsembleNet development, and DetectFormer innovation.

\subsection{Vehicle Detection and Tracking using YOLO and DeepSORT\cite{r3}}

A significant technique combines YOLOv4, known for its cutting-edge object identification capabilities, with DeepSORT for improved tracking efficiency. Using TensorFlow, this method tries to improve traffic management by properly detecting and counting automobiles in video feeds. While YOLOv4 provides a great combination of speed and accuracy, there is a processing performance trade-off that may limit real-time applications on less capable hardware. However, the connection with DeepSORT increases the overall framework for traffic surveillance, making it capable of handling a wide range of vehicle kinds and situations with great precision.

\subsection{EnsembleNet: a hybrid approach for vehicle detection and estimation of traffic density based on faster R-CNN and YOLO models\cite{r4}}
Another novel approach is EnsembleNet, which combines the topologies of Faster R-CNN and YOLOv5. This ensemble approach is intended to improve detection accuracy by using the strengths of both systems. EnsembleNet solves the issues of detecting crowded or small objects by applying data pre-processing, annotation, and augmentation techniques to varied datasets. Despite its increased detection capabilities, the model occasionally issues duplicate detections and heavily relies on careful tuning for maximum performance across varied contexts.

\subsection{DetectFormer: Transformer-Based Object Detection\cite{r5}}
DetectFormer is a big advancement in traffic scene analysis. This transformer-based object detector is designed for autonomous driving systems and includes a Global Extract Encoder (GEE) and a ClassDecoder. DetectFormer distinguishes itself by learning the implicit links between distinct item categories in traffic scenes, resulting in significant gains in object detection accuracy. However, the intricacy of the transformer design and its real-world applicability in actual traffic circumstances offer hurdles for deployment in resource-constrained environments or real-time applications.

\begin{figure*}
\centering
\includegraphics[width=7.0in]{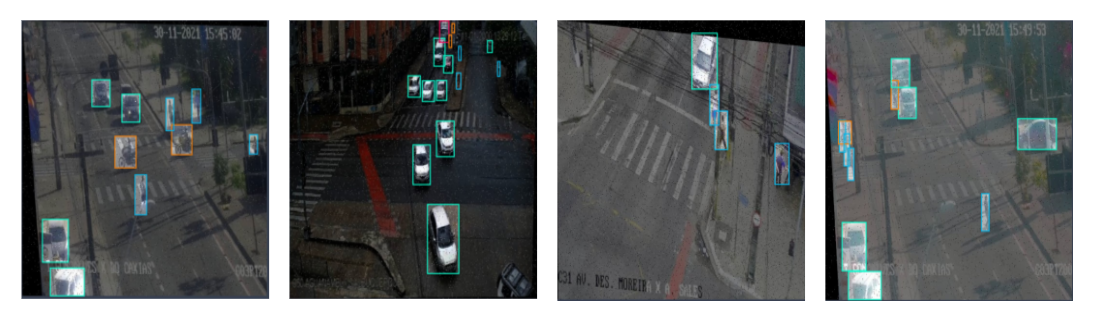}
\caption{Augmented Images from our dataset with Ground Truth bounding boxes.}
\label{scene_example}
\end{figure*}
\section{Data}
\subsection{Data overview}

The data was collected from various public traffic camera websites\cite{r6}. The videos include a wide range of traffic scenarios and vehicle kinds, adding to the thoroughness and dependability of our research approach.
The recordings collected were made both during the day and at night, and they captured a variety of weather conditions. Specific preference was given to videos that capture sidewalks clearly as our goal is to accurately detect pedestrians and small vehicles. The frame rates varied from 25-30 fps at multiple resolutions.


\section{Proposed Framework}
\subsection{Data preprocessing}

\subsubsection{Sampling the frames and Generating the images}
Initially, we acquired some of the raw video data from the different traffic cameras from different periods of time. The traffic video contains day and night videos. Then we sampled all the video frames into images. We took two frames per second to generate the images from the videos. Then we created the dataset by taking 1142 images from different videos.

\subsubsection{Data annotation}
We attempted to autolabel the data. However, the outcomes were insufficient. Next, we categorized all kinds of cars and pedestrians. After the annotation phase, we assembled 1,142 photos to create our baseline dataset. In this collection, 3,388 carefully selected annotations are included, whereas 121 photos are classified as null, meaning they contain no visible objects of interest. The collection includes 2.92 annotations on average per image, covering six different classes. Bicycle (423), bus (163), automobile (998), motorcycle (663), pedestrians (872), and truck (178) are the classes that have the annotations. 

\subsection{Data augmentation}
We utilized 1,142 photos and employed various augmentation techniques to expand the dataset, resulting in a total of 3,082 images. The augmentation methods implemented include:

\noindent\emph{Hue:} It randomly alters the color channels of the input images. It was used to ensure the model was not memorizing a given object or scene's colors. We used hue between -25° and +25°

\noindent\emph{Exposure:} In order to mimic real-world situations, it was utilized to both increase and decrease the exposure of the input photos. We used exposure Between -25\% and +25\%

\noindent\emph{Noise:} A certain percentage of pixels were selected as noise to recreate challenging camera conditions and quality. 5\% noise has been added in the images.

 \noindent\emph{Shear:} For the fallen pedestrian, bicycle and motorcycles we used the sheer. We used  ±15° Horizontal, ±15° Vertical Sheer.

\noindent\emph{Saturation:} Saturation can be detect while there is a low color of the image. It will boost up. We used saturation between -25\% and +25\%.

\noindent\emph{Blur:}Adding controlled blur to images can help make the model more robust to noisy or unclear images. It prevents the model from relying too much on fine details that might be artifacts or irrelevant in real-world scenarios. So we used Up to 2.5px blur. \\

\subsection{Model selection}
We have used two distinct object detection models: YOLOv8, and RT-DETR-L. All of these models are tailored for the single-stage detection of objects. To comprehensively evaluate their performance, multiple variations of each model were employed. 

 \subsubsection{YOLOv8\cite{r2}}
To accommodate a range of visual applications, Ultralytics released five scaled versions of YOLOv8 in January 2023. By combining features and context, the C2f module, which takes the place of CSPLayer, improves accuracy. With a decoupled head and anchor-free model, YOLOv8 achieves better accuracy in objectness, classification, and regression tests. Better bounding box predictions are achieved by the use of DFL losses and CIoU, while the application of binary cross-entropy facilitates the handling of small objects. With support for CLI/PIP usage, YOLOv8-Seg's user-friendliness is further strengthened, and it performs exceptionally well in semantic segmentation.
 
\subsubsection{RT-DETR\cite{r1}}
DETRs, object detectors based on end-to-end transformers, have recently yielded outstanding results. Even with their success, DETRs are difficult to apply because of their high computing needs, which also restrict the use of post-processing techniques such as non-maximum suppression (NMS). To solve this problem, the Real-Time Detection Transformer (RT-DETR) has been presented. In addition to improving query quality and allowing for adaptable speed changes without retraining, RT-DETR manages multi-scale features well. Support for using CLI/PIP further emphasizes how user-friendly it is.

\begin{table}[htbp]
\centering
\caption{Experimental results on validation baseline dataset}
\label{baseline_results}
\begin{tabular}{lccccc}
\hline
Model & mAP & Precision & Recall \\ 
\hline
YOLOv8s & 0.798 & 0.814 & 0.764 \\
\textbf{YOLOv8m} & \textbf{0.898} & \textbf{0.861} & \textbf{0.867}  \\
YOLOv8l & 0.898 & 0.877 & 0.835  \\
YOLOv8x & 0.818 & 0.895 & 0.735  \\
RT-DETR-L & 0.867 & 0.871 & 0.832  \\
RT-DETR-x & 0.878 & 0.886 & 0.85  \\
\hline
\end{tabular}
\end{table}

\begin{table}[htbp]
\centering
\caption{Experimental results on validation augmented dataset}
\label{augmented_results}
\begin{tabular}{lccccc}
\hline
Model & mAP & Precision & Recall \\ 
\hline
YOLOv8s & 0.863 & 0.851 & 0.836 \\
YOLOv8m & 0.889 & 0.86 & 0.842 \\
\textbf{YOLOv8l} & \textbf{0.909} & \textbf{0.884} & \textbf{0.861}  \\
YOLOv8x & 0.842 & 0.872 & 0.755  \\
RT-DETR-L & 0.884 & 0.88 & 0.859  \\
RT-DETR-x & 0.892 & 0.864 & 0.892  \\
\hline
\end{tabular}
\end{table}

\FloatBarrier 

\begin{table}[htbp]
\centering
\caption{Experimental results on Pedestrian Class}
\label{pedestrian_results}
\begin{tabular}{lccccc}
\hline
Model & mAP & Precision & Recall \\ 
\hline
YOLOv8s & 0.778 & 0.879 & 0.717 \\
YOLOv8m & 0.809 & 0.901 & 0.69 \\
\textbf{YOLOv8l} & \textbf{0.822} & \textbf{0.909} & \textbf{0.69} \\
YOLOv8x & 0.726 & 0.884 & 0.687  \\
RT-DETR-L & 0.817 & 0.879 & 0.741  \\
RT-DETR-x & 0.804 & 0.862 & 0.736  \\
\hline
\end{tabular}
\end{table}

\section{Experimental Results}

\begin{figure*}
\centering
\includegraphics[width = 2.2in, height=1.2in]{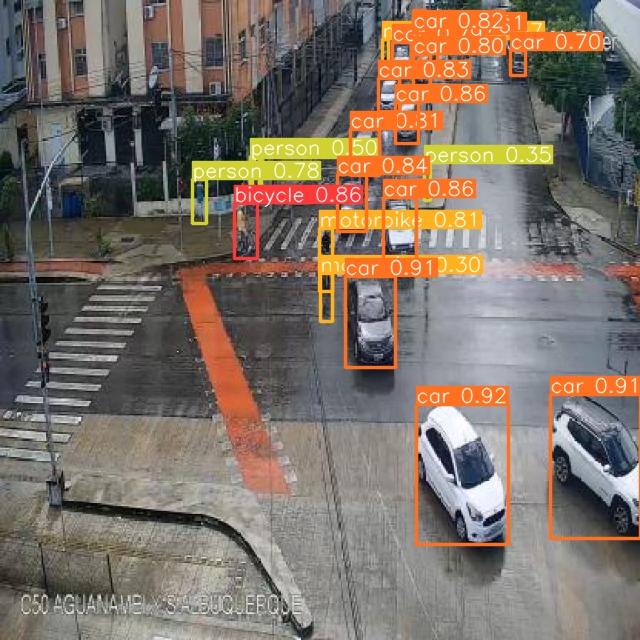}
\includegraphics[width = 2.2in,height=1.2in]{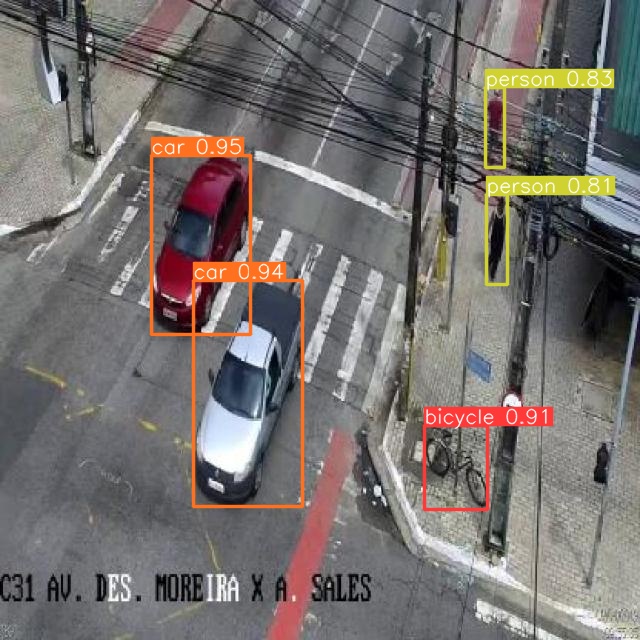}
\includegraphics[width = 2.2in,height=1.2in]{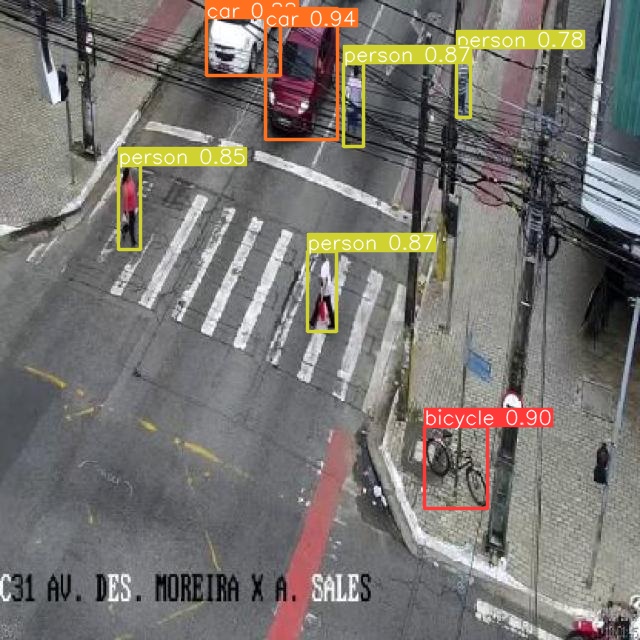}\\
\includegraphics[width = 2.2in, height=1.2in]{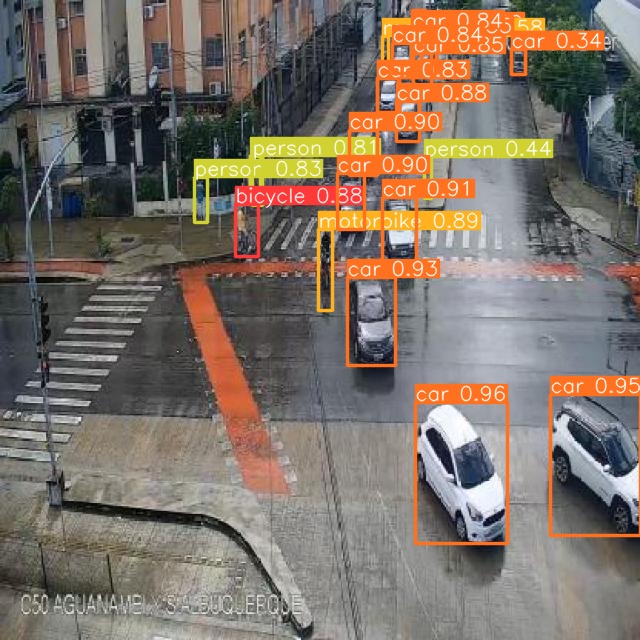}
\includegraphics[width = 2.2in,height=1.2in]{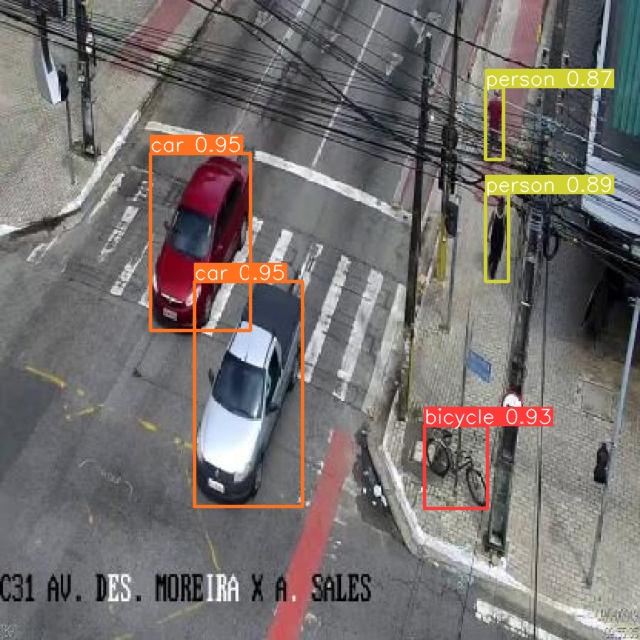}
\includegraphics[width = 2.2in,height=1.2in]{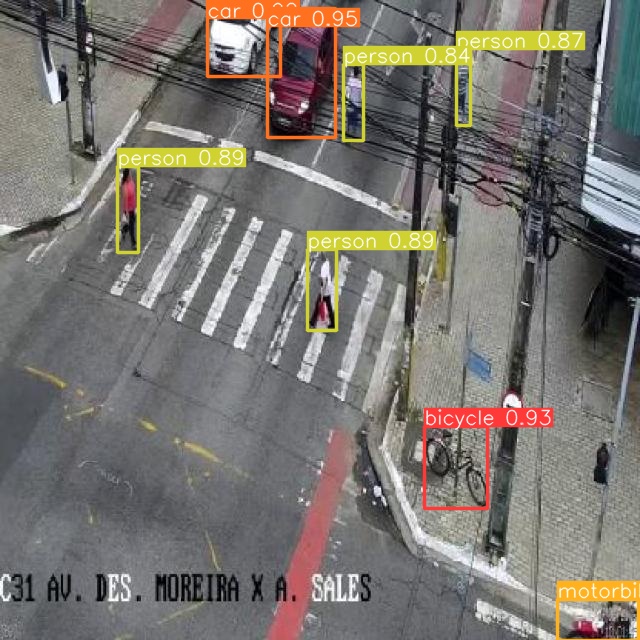}\\
\caption{Row 1: YOLOv8l output images after running the model.Row 2: RT-DETR-L output images after running model.}
\label{example_det}
\end{figure*}

\subsection{Model Training}
During training, the hyperparameter settings remained the same for every model. Every model was trained using a batch size of 8 and 640x640 images for a maximum of 100 epochs. Metrics such as precision, recall, and mean average precision were used to evaluate and contrast the models' effectiveness. The NVIDIA Tesla P1000 GPU, which has 30 GB of RAM and 16 GB of VRAM, was used to power the training and evaluation routines, which were performed on Kaggle.

\subsection{Object Detection Results on baseline dataset}
We conducted a comparative analysis of the models in Table \ref{baseline_results} for the baseline dataset. 
In the evaluation, YOLOv8m emerges as a standout performer, boasting the highest mAP of 0.898, indicative of robust overall performance. Following closely is RT-DETR-x, presenting a competitive mAP of 0.878. YOLOv8s and RT-DETR-L also demonstrate commendable performance, achieving scores of 0.798 and 0.867, respectively. Upon delving into Precision and Recall analysis, YOLOv8m impressively strikes the best balance, achieving Precision and Recall values of 0.861 and 0.867, respectively. YOLOv8l exhibits notable Precision at 0.877, albeit with a slightly lower Recall of 0.835. RT-DETR-x showcases equilibrium with Precision at 0.886 and Recall at 0.85. Notably, YOLOv8x displays a high Precision of 0.895 but a relatively lower Recall of 0.735. Across all metrics, YOLOv8m consistently emerges as a robust performer, making it a noteworthy model in this research evaluation. RT-DETR-x, with competitive results in mAP, Precision, and Recall, underscores its effectiveness in the assessed tasks.

\subsection{Object Detection Results on augmented baseline dataset}

A rigorous examination of diverse models was undertaken in Table \ref{augmented_results}, encompassing an augmented version of our baseline dataset. When compared to other research findings, YOLOv8l performs exceptionally well, with the greatest mean adjusted performance (mAP) of 0.909, indicating strong overall performance after augmentation. Closely trailing is RT-DETR-x, which exhibits a competitive mean approximate performance of 0.892, highlighting its efficacy in the assessed tasks. YOLOv8m also performs admirably, achieving a mAP of 0.889. Examining Precision and Recall, YOLOv8l achieves a balanced Precision of 0.884 and Recall of 0.861, demonstrating proficiency in both categories. Impressively, RT-DETR-x continues to achieve good recall (0.892) and precision (0.864). Furthermore, with Precision and Recall levels of 0.86 and 0.842, respectively, YOLOv8m exhibits a pleasing balance. The table results highlight the unique effectiveness of YOLOv8l on the enhanced dataset, indicating that it is a strong option for situations with a range of conditions. From figure \ref{example_det} we can see the output images for both YOLOv8l and RT-DETR-L to detect and classify the desired objects. 

\subsection{Object Detection Results on pedestrian class}

Specifically, the pedestrian class was undertaken in Table \ref{pedestrian_results} this result is based on the augmented dataset. In our study, YOLOv8l achieves the greatest mean accuracy point (mAP) of 0.822, indicating strong performance in pedestrian detection and categorization. Closely trailing is RT-DETR-L, which has a competitive mAP of 0.817, indicating that it performs well in the tasks that are assessed. YOLOv8m also performs admirably, achieving a mAP of 0.809. A careful examination of Precision and Recall reveals that YOLOv8l has the greatest Precision of 0.909, highlighting its remarkable accuracy in detecting pedestrians. Both RT-DETR-L and YOLOv8s exhibit an excellent trade-off between Precision and Recall, which enhances their efficacy. Interestingly, YOLOv8x has a little lower Recall (0.687) but a higher Precision (0.884). The best performer is YOLOv8l, which demonstrates how well it can distinguish and classify pedestrians. Competitive results are also presented by RT-DETR-L and YOLOv8m, which show a nice balance between Precision and Recall.

YOLOv8 also performed better than RT-DETR in terms of speed. On the test set YOLOv8 was doing inference at 40 FPS while RT-DETR was doing 25 FPS. As most traffic cameras operate at 24 FPS, both these models can be used for real-time applications. 

\section{Conclusion}
We have successfully developed a sophisticated system for multi-class traffic analysis in this study, showcasing a wide range of capabilities. Under various urban contexts and conditions, our system performs exceptionally well in terms of item detection, classification. The method consists of a meticulously designed two-step procedure that begins with significant frame extraction and integrates single-stage object detection with conventional post-processing and data analysis methods. We have achieved compelling results through extensive experimentation, demonstrating the remarkable efficacy and resilience of our approach in vehicle analysis. In summary, our built system is a great asset in the field of traffic monitoring and management because it not only performs extraordinarily well at the moment but also has the potential to continue improving in the future.

\bibliography{conference_101719}{}
\bibliographystyle{plain}
\end{document}